\documentclass[11pt]{article}
\usepackage{silence}
\newif\ifanonymous
\anonymousfalse

\ifanonymous
  \usepackage[review]{acl}   
\else
  \usepackage{acl}           
\fi

\usepackage{times}
\usepackage{latexsym}

\usepackage[T1]{fontenc}
\usepackage{inconsolata}

\usepackage{graphicx}
\usepackage{booktabs}
\usepackage{multirow}
\usepackage{amsmath,amssymb}
\usepackage{url}

\usepackage{tikz}
\usetikzlibrary{positioning,arrows.meta,shapes,fit}
\usepackage{pgfplots}
\pgfplotsset{compat=1.18}
\usepackage{tikz}
\usetikzlibrary{positioning,calc,arrows.meta}
\usepackage{placeins} 
\usepackage{algorithm}
\usepackage{algorithmic}
\usepackage{microtype}

\title{CRANE: Causal Relevance Analysis of Language-Specific Neurons in Multilingual Large Language Models}

\ifanonymous
\author{
  Anonymous Authors \\
  Affiliation \\
  \texttt{email@domain}
}
\else
\author{
  Yifan Le$^{1}$\thanks{Corresponding author.} \quad
  Yunliang Li$^{1}$ \\
  $^{1}$Zhejiang University \\
  \texttt{\{leyifan, liyunliang\}@zju.edu.cn} 
}
\fi

\begin{document}
\maketitle

\begin{abstract}
Multilingual large language models (LLMs) achieve strong performance across languages, yet how language capabilities are organized at the neuron level remains poorly understood.
Prior work has identified language-related neurons mainly through activation-based heuristics, which conflate language preference with functional importance.
We propose \textbf{CRANE}, a relevance-based analysis framework that \emph{redefines language specificity in terms of functional necessity}, identifying language-specific neurons through targeted neuron-level interventions.
CRANE characterizes neuron specialization by their contribution to language-conditioned predictions rather than activation magnitude.
Our implementation will be made publicly available.
Neuron-level interventions reveal a consistent asymmetric pattern: masking neurons relevant to a target language selectively degrades performance on that language while preserving performance on other languages to a substantial extent, indicating language-selective but non-exclusive neuron specializations.
Experiments on English, Chinese, and Vietnamese across multiple benchmarks, together with a dedicated relevance-based metric and base-to-chat model transfer analysis, show that CRANE isolates language-specific components more precisely than activation-based methods.
\end{abstract}

\section{Introduction}

Large Language Models (LLMs) have achieved remarkable progress in multilingual natural language processing, enabling a single model to perform understanding and generation across many languages.
Foundational multilingual models such as Meta’s LLaMA2 \citep{touvron2023llama2}, Google DeepMind’s Gemini~1.5 \citep{gemini1p5}, and emerging open-source GPT families such as GPT-OSS \citep{agarwal2025gpt} have demonstrated strong performance across diverse linguistic settings.
Despite these empirical advances, how language capabilities are organized and specialized at the neuron level remains poorly understood.

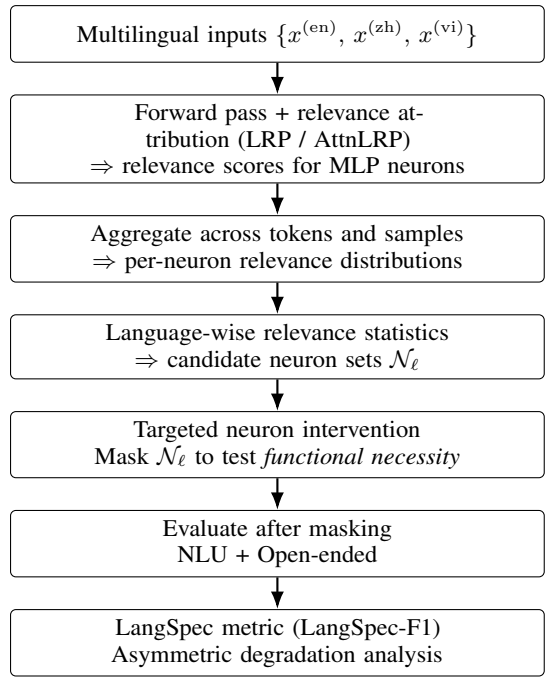
\begin{figure}[t]
\centering
\begin{tikzpicture}[
  font=\small,
  node distance=4.2mm,
  box/.style={
    draw,
    rounded corners=2pt,
    align=center,
    inner sep=3.5pt,
    minimum height=7.5mm,
    text width=0.88\linewidth
  },
  arrow/.style={-Latex, line width=0.8pt}
]

\node[box] (input)
{Multilingual inputs $\{x^{(\mathrm{en})},\,x^{(\mathrm{zh})},\,x^{(\mathrm{vi})}\}$};

\node[box, below=of input] (attrib)
{Forward pass + relevance attribution (LRP / AttnLRP)\\
$\Rightarrow$ relevance scores for MLP neurons};

\node[box, below=of attrib] (dist)
{Aggregate across tokens and samples\\
$\Rightarrow$ per-neuron relevance distributions};

\node[box, below=of dist] (select)
{Language-wise relevance statistics\\
$\Rightarrow$ candidate neuron sets $\mathcal{N}_\ell$};

\node[box, below=of select] (intervene)
{Targeted neuron intervention\\
Mask $\mathcal{N}_\ell$ to test \emph{functional necessity}};

\node[box, below=of intervene] (eval)
{Evaluate after masking\\
NLU + Open-ended};

\node[box, below=of eval] (metric)
{LangSpec metric (LangSpec-F1)\\
Asymmetric degradation analysis};

\draw[arrow] (input) -- (attrib);
\draw[arrow] (attrib) -- (dist);
\draw[arrow] (dist) -- (select);
\draw[arrow] (select) -- (intervene);
\draw[arrow] (intervene) -- (eval);
\draw[arrow] (eval) -- (metric);

\end{tikzpicture}

\vspace{-2mm}
\caption{\textbf{Overview of CRANE.} CRANE identifies language-specific neurons via relevance attribution and language-wise relevance statistics, and validates \emph{functional necessity} through targeted neuron-level interventions with unified evaluation across NLU and open-ended benchmarks.}
\label{fig:crane_overview}
\vspace{-2mm}
\end{figure}

A growing line of work has attempted to identify \emph{language-specific neurons}, often based on activation statistics or language-conditioned probing.
Earlier studies on recurrent neural networks examined gated units as mechanisms for information storage and control \citep{hochreiter1997lstm,elman1990finding}, while later work analyzed intrinsic linguistic knowledge in sequential models \citep{qian-etal-2016-analyzing}.
More recently, neuron-level analyses in LLMs have focused on activation distributions and activation probabilities \citep{tang2024language_specific_neurons}, as well as finer-grained neuron identification approaches \citep{dai-etal-2022-knowledge, zhang2025alignment_language_neurons}.
These studies reveal statistical regularities associated with language, but largely assume that statistical correlation reflects functional importance.

Activation does not imply functional necessity.
Prior work identifies language-specific neurons based on activation statistics, but typically lacks direct functional validation \citep{tang2024language_specific_neurons}.
Consequently, neurons correlated with a language may not be required for its performance.
We instead define language specificity in terms of \emph{functional necessity}.

We propose \textbf{CRANE} (\textit{Causal Relevance-based Analysis of Neuron Specialization}), a relevance-based framework that operationalizes this definition through targeted neuron-level interventions.
CRANE builds on layer-wise relevance propagation (LRP) \citep{arras2016what_is_relevant,arras2017explaining_rnn} and its Transformer extension AttnLRP \citep{pmlr-v235-achtibat24a} to attribute language-conditioned predictions to individual neurons.
By identifying neurons based on output-level relevance rather than activation magnitude, CRANE distinguishes language preference from functional contribution.

Using CRANE, we uncover a consistent asymmetric specialization pattern across languages.
Masking neurons relevant to a target language leads to substantially larger degradation on that language while generally preserving performance on others.
This pattern supports \emph{language-selective but non-exclusive} specialization: neurons contribute disproportionately to specific languages while remaining part of shared multilingual computation.
We validate these findings on English, Chinese, and Vietnamese across multiple multilingual benchmarks.

In summary, this work makes four main contributions.

(1) We redefine language specificity at the neuron level by shifting from
activation-based correlation to functional necessity, and provide a concrete
operationalization via CRANE.

(2) We introduce a relevance-based evaluation metric that quantifies
language-selective functional effects under targeted neuron intervention,
enabling systematic comparison across languages and models.

(3) We present functional evidence of asymmetric, non-exclusive language
specialization in multilingual LLMs, where neurons contribute
disproportionately to specific languages while remaining involved in
multilingual computation.

(4) We conduct a controlled transfer analysis from pretrained Base models
to post-trained Chat models without re-identification, providing empirical
insight into how language-selective neuron effects persist or shift after
instruction tuning.

\begin{figure}[t]
  \centering
  \includegraphics[width=\columnwidth]{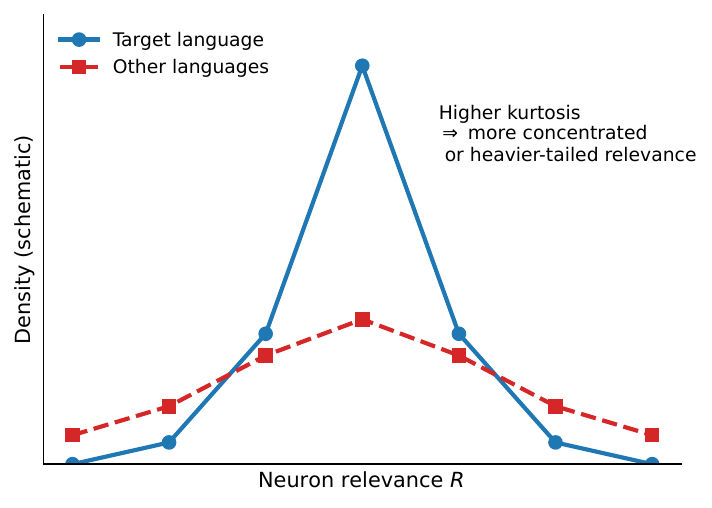}
  \caption{Schematic intuition of kurtosis-based language contrast.
  Under a target language condition, a neuron may exhibit a more concentrated
  or heavy-tailed relevance distribution than under other languages.
  Kurtosis is used as one example statistic to quantify such concentration differences.}
  \label{fig:kurtosis_intuition}
\end{figure}

\section{Related Work}

\paragraph{Correlation-based explanations for LLMs.}
Many interpretability methods for LLMs extract correlational explanatory signals.
Feature attribution techniques such as Integrated Gradients~\cite{sundararajan2017axiomatic} and perturbation-based approaches~\cite{li2016visualizing} estimate input contributions, while attention-based visualizations~\cite{vig2019bertviz} provide intuitive but potentially unfaithful explanations~\cite{jain2019attention}.
Probing-based analyses~\cite{belinkov2022probing} and neuron- or concept-level studies~\cite{dalvi2019one,kim2018interpretability} reveal representational patterns but may conflate selectivity with importance~\cite{hewitt2019designing}.
Overall, these approaches characterize statistical or descriptive signals, without establishing functional necessity through intervention.

\paragraph{Multilingual structure and language-selective components.}
For multilingual models, prior work studies cross-lingual alignment and shared representations using probing and representation similarity analysis~\cite{karthikeyan2020mbert,blevins2022pretraining_dynamics}. 
More recent studies investigate language-specific or language-selective neurons by analyzing activation statistics or frequency, exemplified by LAPE~\cite{tang2024language_specific_neurons} and related neuron-level methods~\cite{zhang2025alignment_language_neurons}. 
While these methods uncover language-related structure, they typically equate activation selectivity with functional importance. 
As noted in prior work, such selectivity does not guarantee that intervening on the identified neurons induces targeted language degradation~\cite{tang2024language_specific_neurons}.

\paragraph{Intervention-based and causal analyses.}
A growing line of mechanistic interpretability emphasizes causal validation via intervention.
Methods such as activation patching and causal tracing intervene on internal activations to localize computations responsible for model behavior~\cite{meng2022rome,zhang2024activation_patching_best_practices}.
These approaches motivate defining interpretability in terms of functional necessity rather than correlational signals.
Our work aligns with this perspective and applies intervention-based validation to study language specialization in multilingual LLMs.


\section{Method}

CRANE operationalizes \emph{functional necessity} as the definition of language specificity at the neuron level.
Rather than inferring language specificity from activation-based correlation, CRANE evaluates whether intervening on a neuron set induces a disproportionately larger performance degradation on a target language than on others.
This definition-driven criterion guides all components of the framework, from neuron attribution to intervention-based validation.
Figure~\ref{fig:crane_overview} illustrates the overall workflow.

\subsection{Problem Setup and Neuron-level Representation}

Let the input token sequence be $\mathbf{x}=\{x_i\}_{i=1}^{N}$ and the model output logits be $\mathbf{y}\in\mathbb{R}^{|V|}$, where $|V|$ denotes the vocabulary size.
For layer $l$, the hidden state at token position $i$ is $\mathbf{h}^{(l)}_i\in\mathbb{R}^{d}$.

We analyze standard Transformer architectures without architectural modification.
Attention layers are treated as contextual mixing modules, while neuron-level analysis is conducted on MLP components, where each \textbf{column} of the linear projection matrices is treated as an individual neuron, consistent with prior neuron-level analyses.
Given a set of languages $\mathcal{L}$, our goal is to identify, for each language $\ell\in\mathcal{L}$, a neuron subset $\mathcal{N}_{\ell}$ such that intervening on $\mathcal{N}_{\ell}$ results in a larger relative performance degradation on language $\ell$ than on non-target languages under the same intervention budget.

\begin{figure}[t]
    \centering
    \includegraphics[width=\linewidth]{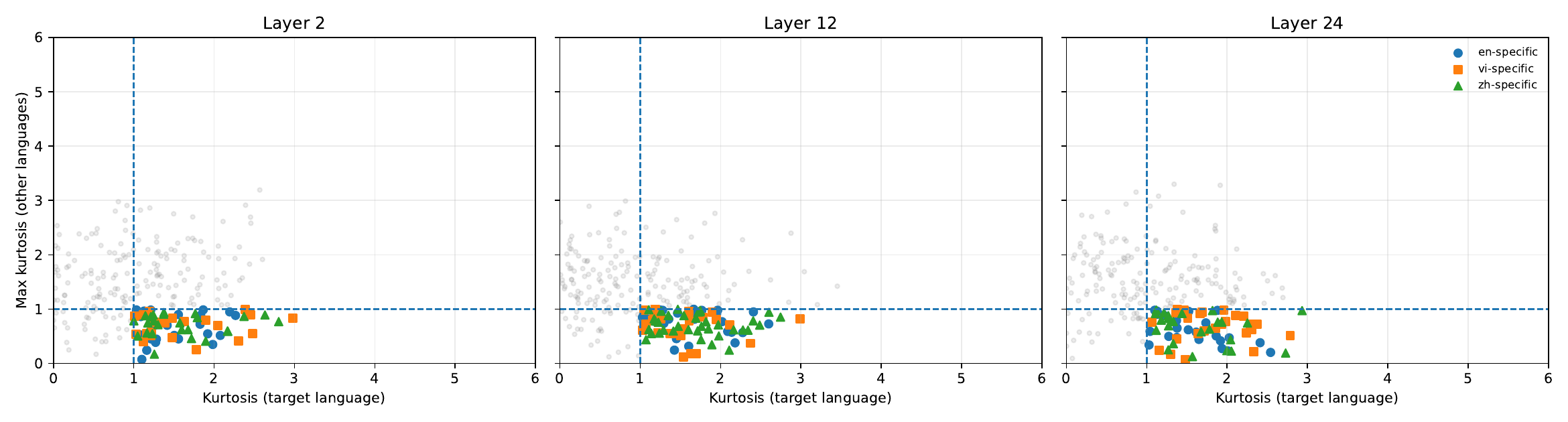}
    \caption{
    Kurtosis-based contrast for identifying language-specific neurons.
    Each subplot corresponds to one Transformer layer.
    Each point represents a neuron, plotted by its normalized kurtosis under a target language
    (x-axis) versus the maximum kurtosis under other languages (y-axis).
    Dashed lines indicate the kurtosis threshold (set to 1).
    Neurons with high kurtosis for the target language but low kurtosis for others
    (bottom-right region) exhibit stronger language-specific relevance concentration.
    }
    \label{fig:kurtosis-contrast}
\end{figure}

\subsection{Neuron-level Relevance Attribution}

To estimate the contribution of individual neurons to model outputs, CRANE employs \emph{layer-wise relevance propagation} (LRP), a relevance attribution technique that redistributes output relevance backward through the network while preserving relevance conservation \citep{arras2016what_is_relevant,arras2017explaining_rnn}.
We adopt existing extensions of LRP for Transformer architectures \citep{pmlr-v235-achtibat24a}.

Given an input $\mathbf{x}$ in language $\ell$ and a language-conditioned output objective $f_{\ell}(\mathbf{x})$, relevance is initialized at the output layer and propagated backward according to the conservation principle:
\begin{equation}
\sum R^{(l)} = \sum R^{(l-1)}.
\end{equation}

Relevance is propagated to MLP neurons, yielding token-level relevance scores.
These scores are aggregated across tokens to obtain a sample-level relevance value per neuron.
We emphasize that relevance attribution serves as a \emph{tool} for estimating neuron contributions within CRANE, rather than constituting the method itself.

\subsection{Language-conditioned Relevance Distributions}

CRANE characterizes language specificity at the \emph{distributional} level across samples.
For each neuron $n$ and language $\ell$, aggregating relevance values over a large set of inputs induces a language-conditioned relevance distribution.

Intuitively, neurons that are functionally necessary for a target language tend to exhibit more concentrated or heavy-tailed relevance distributions under that language compared to others.
Figure~\ref{fig:kurtosis_intuition} provides a schematic illustration of this intuition.
To quantify such concentration differences, we adopt kurtosis, a fourth-order moment statistic that captures distributional peakedness and tail behavior \citep{decarlo1997kurtosis}.

Formally, the kurtosis of a neuron's relevance distribution $R_n$ is defined as:
\begin{equation}
\mathrm{kurtosis}(R_n)=\frac{\mathbb{E}[(R_n-\mu)^4]}{\sigma^4},
\end{equation}
where $\mu$ and $\sigma$ denote the mean and standard deviation, respectively.

We compare language-conditioned kurtosis values across languages and rank neurons by their relative concentration under the target language.
In practice, kurtosis scores are normalized within each layer, and neurons whose normalized scores exceed a fixed threshold are selected to form candidate sets.
This threshold is used to control the intervention budget rather than to define language specificity itself.
We show in experiments that the resulting findings are robust across a reasonable range of thresholds.

\subsection{Neuron-level Intervention and Validation}

Given a candidate neuron set $\mathcal{N}_{\ell}$, CRANE validates functional necessity through targeted neuron-level intervention.
Specifically, neurons in $\mathcal{N}_{\ell}$ are masked by setting their outputs to zero during inference, while all other components remain unchanged.

Rather than assuming strict language exclusivity, we evaluate whether masking $\mathcal{N}_{\ell}$ induces a larger relative performance degradation on the target language than on non-target languages under identical intervention budgets.
This relative degradation criterion provides functional evidence for \emph{language-selective but non-exclusive} neuron specialization.

We evaluate intervention effects on both natural language understanding (NLU) and open-ended generation benchmarks.
The complete CRANE procedure is summarized in Algorithm~\ref{alg:crane}.

\begin{figure}[t]
  \centering
  \includegraphics[width=\columnwidth]{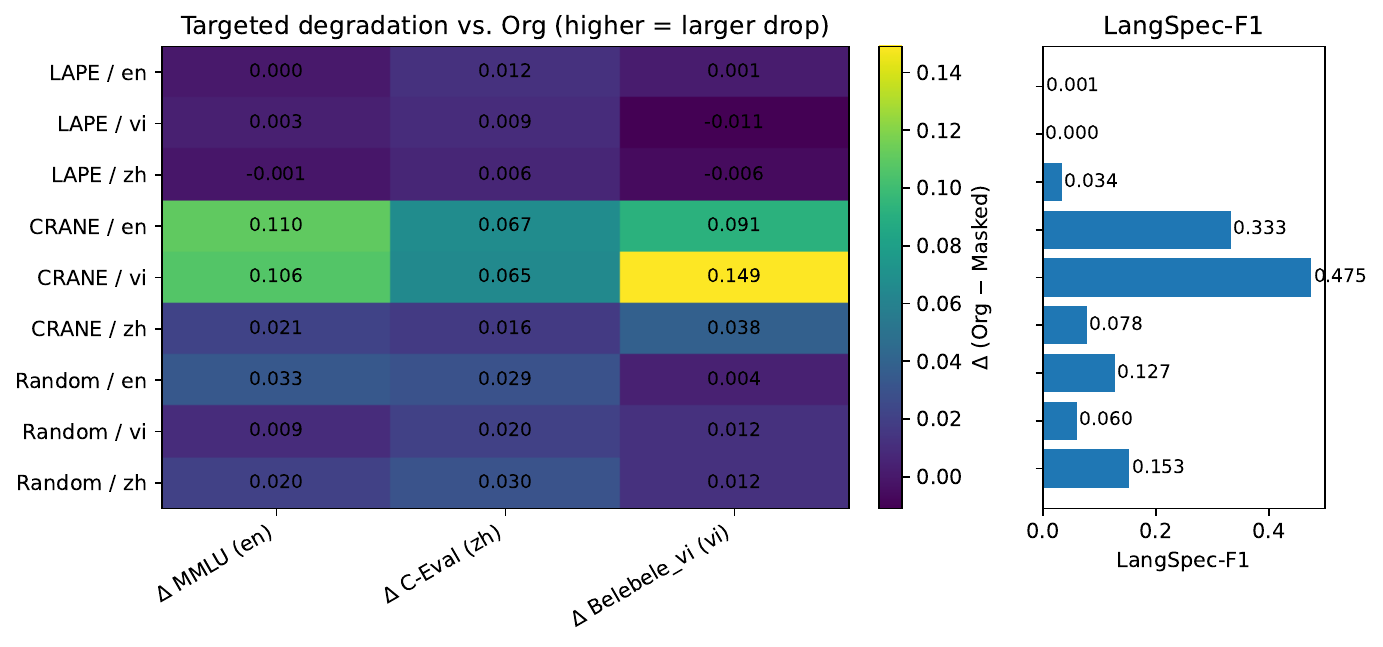}
  \caption{
Targeted NLU degradation under neuron-level intervention on \textsc{LLaMA2-7B-Base}.
The heatmap reports absolute performance drops ($\Delta = \text{Org} - \text{Masked}$)
on each evaluation benchmark (columns) when masking neuron sets identified for a
given target language and method (rows).
Higher values indicate stronger degradation.
The bar plot summarizes language-specific functional effects using LangSpec-F1.
}

  \label{fig:nlu_drop_heatmap}
\end{figure}

\subsection{Language-specificity Metric}
\label{sec:langspec_metric}
To quantify language-specific functional necessity under neuron-level intervention, we introduce \textbf{LangSpec-F1}, a composite metric that balances targeted performance degradation on the masked language with stability on non-target languages. LangSpec-F1 is a metric to quantify functional selectivity, not strict exclusivity; high LangSpec-F1 indicates that neuron interventions disproportionately affect the target language relative to others.

Let $\Delta_{\ell}$ denote the performance drop on the target language $\ell$ after masking a neuron set, and let $\max_{\ell' \neq \ell} \Delta_{\ell'}$ denote the maximum performance drop observed on any non-target language.
We consider only negative performance changes by defining $\Delta = \max(\mathrm{drop}, 0)$, such that performance improvements are not counted as degradation.

We define precision and recall as:
\begin{equation}
\mathrm{Precision} = \frac{\Delta_{\ell}}{\Delta_{\ell} + \max_{\ell' \neq \ell} \Delta_{\ell'} + \epsilon},
\end{equation}
\begin{equation}
\mathrm{Recall} = \frac{\Delta_{\ell}}{S_{\ell} + \epsilon},
\end{equation}
where $S_{\ell}$ denotes the original (unmasked) performance score on language $\ell$, and $\epsilon$ is a small constant for numerical stability.

The final LangSpec-F1 score is computed as the harmonic mean of precision and recall:
\begin{equation}
\mathrm{LangSpec\text{-}F1} = 
\frac{2 \cdot \mathrm{Precision} \cdot \mathrm{Recall}}
{\mathrm{Precision} + \mathrm{Recall} + \epsilon}.
\end{equation}

We emphasize that LangSpec-F1 is an operational metric for evaluating functional necessity, and that these design choices reflect a conservative assessment of non-target language interference rather than a definition of language specificity itself.

Intuitively, LangSpec-F1 assigns high scores to neuron sets whose intervention causes substantial degradation on the target language while inducing minimal degradation on non-target languages, relative to the original performance level.

\begin{table*}[t]
\centering
\small
\begin{tabular}{l l c c c c}
\toprule
\textbf{Method} & \textbf{Mask lang} & \textbf{MMLU (en)} & \textbf{C-Eval (zh)} & \textbf{Belebele\_vi (vi)} & \textbf{LangSpec-F1} \\
\midrule
Org    & -- & 0.4579 & 0.3470 & 0.3722 & -- \\
\midrule
LAPE   & en & 0.4576 & 0.3351 & 0.3711 & 0.0013 \\
LAPE   & vi & 0.4553 & 0.3380 & 0.3833 & 0.0000 \\
LAPE   & zh & 0.4589 & 0.3410 & 0.3778 & 0.0337 \\
\midrule
CRANE  & en & \textbf{0.3483} & \textbf{0.2801} & \textbf{0.2811} & \textbf{0.3328} \\
CRANE  & vi & \textbf{0.3517} & \textbf{0.2816} & \textbf{0.2233} & \textbf{0.4747} \\
CRANE  & zh & 0.4366 & 0.3314 & 0.3344 & 0.0779 \\
\midrule
Random & en & 0.4249 & 0.3180 & 0.3678 & 0.1270 \\
Random & vi & 0.4486 & 0.3269 & 0.3600 & 0.0603 \\
Random & zh & 0.4375 & 0.3165 & 0.3600 & 0.1531 \\
\bottomrule
\end{tabular}
\caption{NLU results on LLaMA2-7B-Base with neuron masking. Higher is better for task metrics. LangSpec-F1 reflects targeted degradation and non-target stability.}
\label{tab:nlu}
\end{table*}

\section{Experimental Setup}

\subsection{Languages and Models}

We study three typologically diverse languages: English (en), Chinese (zh), and Vietnamese (vi).
Experiments are conducted at two levels.
First, we identify and validate language-specific neuron sets on \textbf{LLaMA2-7B-Base}.
Second, we transfer neuron sets discovered on the base model to \textbf{LLaMA2-7B-Chat} to examine whether language-specific functional necessity is preserved after post-training.
This setting allows us to probe the stability of neuron-level specialization across pretraining and post-training stages.

\subsection{Baselines}

We compare CRANE against the following baselines, all evaluated under identical intervention budgets:

\begin{itemize}
  \item \textbf{LAPE}~\cite{tang2024language_specific_neurons}: an activation-based approach that identifies language-related neurons by measuring neuron activation likelihood across language-specific corpora.
  \item \textbf{Random masking (budget-matched)}: randomly sampling the same number of neurons from identical module structures, serving as a control to distinguish targeted language effects from generic perturbations.
\end{itemize}

\subsection{Tasks and Evaluation}

\paragraph{Natural language understanding (NLU).}
We evaluate language understanding using standard language-specific benchmarks:
\textbf{MMLU} for English~\citep{hendrycks2021mmlu},
\textbf{C-Eval} for Chinese~\citep{huang2023ceval},
and \textbf{Belebele} for Vietnamese~\citep{bandarkar2023belebele}.
All NLU evaluations are conducted using a unified evaluation framework, \texttt{lm-evaluation-harness}~\citep{biderman2024lm_eval_harness}, with consistent prompting and decoding settings across languages and methods.

\paragraph{Open-ended generation.}
We evaluate open-ended generation using an English/Chinese/Vietnamese question set following the LAPE protocol.
Model outputs are scored on a 1--10 scale by an LLM-as-a-judge based on GPT-4o\citep{hurst2024gpt4o}.
The same evaluation protocol is applied before and after neuron-level intervention to ensure direct comparability.

\paragraph{Language-specificity metric.}
We report \textbf{LangSpec-F1}, a composite metric that captures
(i) performance degradation on the target language after neuron-level intervention and
(ii) relative performance retention on non-target languages under the same intervention budget.
The formal definition and computation of LangSpec-F1 are provided in Section~\ref{sec:langspec_metric}.

\subsection{Inference and Implementation Details}

Inference and evaluation are performed using \textbf{vLLM}\citep{kwon2023vllm}.
We use greedy decoding with repetition\_penalty=$1.1$, max\_new\_tokens=$2048$, and a maximum context length of $8192$.
For each identified neuron set, we evaluate all tasks both \textbf{before} and \textbf{after} masking to ensure direct comparison of intervention effects.

\section{Results}

We report results on \textsc{LLaMA2-7B-Base} and \textsc{LLaMA2-7B-Chat}.
We analyze distributional relevance statistics to characterize language-related structure,
validate functional effects via neuron-level intervention on NLU and open-ended generation,
and examine transferability from the Base model to the Chat model.

\begin{table*}[t]
\centering
\small
\begin{tabular}{l l c c c c}
\toprule
\textbf{Method} & \textbf{Mask lang} & \textbf{Open\_en} & \textbf{Open\_vi} & \textbf{Open\_zh} & \textbf{LangSpec-F1} \\
\midrule
Org    & -- & 3.2286 & 1.9286 & 2.3714 & -- \\
\midrule
LAPE   & en & 3.0429 & 2.0143 & 2.2857 & 0.1061 \\
LAPE   & vi & 3.3286 & 2.2571 & 2.2429 & 0.0000 \\
LAPE   & zh & 3.0000 & 2.1714 & 2.6714 & 0.0000 \\
\midrule
CRANE  & en & \textbf{1.6429} & 1.5286 & \textbf{1.2429} & \textbf{0.5337} \\
CRANE  & vi & 2.1286 & \textbf{1.7143} & 1.7000 & 0.1322 \\
CRANE  & zh & 2.5714 & 1.5857 & \textbf{1.4714} & \textbf{0.4582} \\
\midrule
Random & en & 2.1000 & 1.8714 & 1.7286 & 0.4406 \\
Random & vi & 2.7286 & 1.8571 & 1.4286 & 0.0751 \\
Random & zh & 2.5857 & 2.0143 & 2.1571 & 0.1422 \\
\bottomrule
\end{tabular}
\caption{Open-ended generation results on LLaMA2-7B-Base (LLM-judge scores; higher is better).}
\label{tab:open-base}
\end{table*}

\subsection{Normalized Kurtosis Reveals Language-related Structure}

After computing neuron-level relevance scores, we analyze the \emph{normalized kurtosis}
of relevance distributions under different language conditions.
For each neuron, we compare its language-conditioned kurtosis values to assess how strongly
its relevance concentrates for one language relative to others.

We observe clear separation patterns across languages, indicating that normalized kurtosis
captures systematic differences in how neurons participate in computation under different
language inputs.
Importantly, this analysis reflects \emph{language-related structure} at the distributional level
rather than functional language specificity.
Accordingly, kurtosis-based selection serves solely as a candidate identification step,
with functional necessity established only through subsequent intervention.

Figure~\ref{fig:kurtosis-contrast} visualizes this structure by plotting neurons in a
two-dimensional contrast space defined by target-language versus non-target normalized kurtosis.
Neurons exhibiting high target-language kurtosis and low non-target kurtosis form a distinct region,
motivating our candidate selection criterion.

\subsection{NLU: CRANE Induces Targeted Functional Degradation}

Table~\ref{tab:nlu} reports NLU results on \textsc{LLaMA2-7B-Base} under different neuron masking strategies.
Activation-based baselines such as LAPE induce small and relatively uniform performance changes across
languages, yielding LangSpec-F1 values close to zero.
In contrast, neuron sets selected by CRANE produce substantially larger degradation on the intended
target language under identical intervention budgets.

For example, when targeting Vietnamese, \textsc{Belebele\_vi} accuracy decreases from 0.3722
(unmasked) to 0.2233 after masking CRANE-selected neurons, yielding a LangSpec-F1 of 0.4747.
Across all language and benchmark settings, CRANE consistently yields higher LangSpec-F1 than
activation-based and random baselines, indicating robust target-aligned degradation while preserving
non-target performance.

Figure~\ref{fig:nlu_drop_heatmap} summarizes absolute performance drops on NLU benchmarks relative
to the unmasked model.
Compared to baselines, CRANE yields larger drops on target-language benchmarks and smaller changes
on non-target languages, consistent with language-selective but non-exclusive functional effects.

\begin{algorithm}[t]
\caption{CRANE: Causal Relevance-based Analysis of Neuron Specialization}
\label{alg:crane}
\small
\begin{algorithmic}[1]
\REQUIRE Multilingual dataset $\mathcal{D}=\{\mathcal{D}_\ell\}_{\ell\in\mathcal{L}}$, model $f$, languages $\mathcal{L}$, intervention budget $B$
\ENSURE Candidate neuron sets $\{\mathcal{N}_\ell\}_{\ell\in\mathcal{L}}$ and post-intervention evaluation metrics

\STATE \textbf{// Stage 1: Neuron relevance attribution}
\FOR{each language $\ell \in \mathcal{L}$}
    \FOR{each sample $x \in \mathcal{D}_\ell$}
        \STATE Run forward pass and compute objective $s=f(x)$
        \STATE Apply LRP/AttnLRP to propagate relevance to MLP neurons
        \STATE Obtain neuron relevance vector $\mathbf{r}(x,\ell)\in\mathbb{R}^{N}$ \COMMENT{$N$: number of MLP neurons}
    \ENDFOR
    \STATE Aggregate $\{\mathbf{r}(x,\ell)\}_{x\in\mathcal{D}_\ell}$ to form per-neuron relevance distributions $\{\mathcal{R}_{n,\ell}\}_{n=1}^{N}$
\ENDFOR

\STATE \textbf{// Stage 2: Language-conditioned relevance concentration}
\FOR{each language $\ell \in \mathcal{L}$}
    \FOR{each neuron $n=1,\dots,N$}
        \STATE Compute kurtosis $\kappa_{n,\ell}$ of $\mathcal{R}_{n,\ell}$
    \ENDFOR
    \STATE Normalize $\{\kappa_{n,\ell}\}_{n=1}^{N}$ within each layer to obtain normalized scores $\{\tilde{\kappa}_{n,\ell}\}$
\ENDFOR

\STATE \textbf{// Stage 3: Candidate neuron selection (budget control)}
\FOR{each target language $\ell \in \mathcal{L}$}
    \STATE Select $\mathcal{N}_\ell \leftarrow \textsc{BudgetSelect}(\{\tilde{\kappa}_{n,\ell}\}_{n=1}^{N}, B)$
    \COMMENT{e.g., threshold on $\tilde{\kappa}$ or Top-$B$ ranking}
\ENDFOR

\STATE \textbf{// Stage 4: Functional necessity validation (intervention)}
\FOR{each target language $\ell \in \mathcal{L}$}
    \STATE Mask neuron outputs for $\mathcal{N}_\ell$ (set outputs to $0$) during inference
    \STATE Evaluate NLU and open-ended benchmarks across languages under the same budget $B$
    \STATE Compute LangSpec-F1 from relative target-language degradation vs.\ non-target retention
\ENDFOR

\RETURN $\{\mathcal{N}_\ell\}_{\ell\in\mathcal{L}}$ and evaluation metrics
\end{algorithmic}
\end{algorithm}

\subsection{Open-ended Generation: Base Model}

Table~\ref{tab:open-base} reports open-ended generation results on \textsc{LLaMA2-7B-Base}.
Activation-based baselines yield weak or inconsistent intervention effects,
whereas CRANE induces clearer target-language degradation and achieves higher LangSpec-F1
in multiple cases.
Given the inherent variability of open-ended evaluation, these results are presented as
supportive evidence of functional influence.

\subsection{Transferability of Base-identified Neurons to the Chat Model}
\label{sec:transfer}

We examine whether neuron sets identified on the pretrained Base model retain functional
influence after post-training.
Under a strict transfer setting, neuron sets are identified only on
\textsc{LLaMA2-7B-Base} and directly transferred to \textsc{LLaMA2-7B-Chat} without re-identification.
Observed target-language degradation under this setting indicates that a subset of
Base-identified neurons continues to contribute functionally after post-training.

\paragraph{NLU transfer.}
Table~\ref{tab:transfer_nlu} reports NLU results under transferred neuron masks.
Activation-based baselines induce small and inconsistent changes,
whereas CRANE-selected neuron sets produce larger degradation on the target language
on the Chat model, yielding higher LangSpec-F1 in several settings, most notably for Vietnamese.
These findings indicate partial preservation of functional influence rather than invariance
of neuron identity across training stages.

\begin{figure*}[t]
  \centering
  \includegraphics[width=\textwidth]{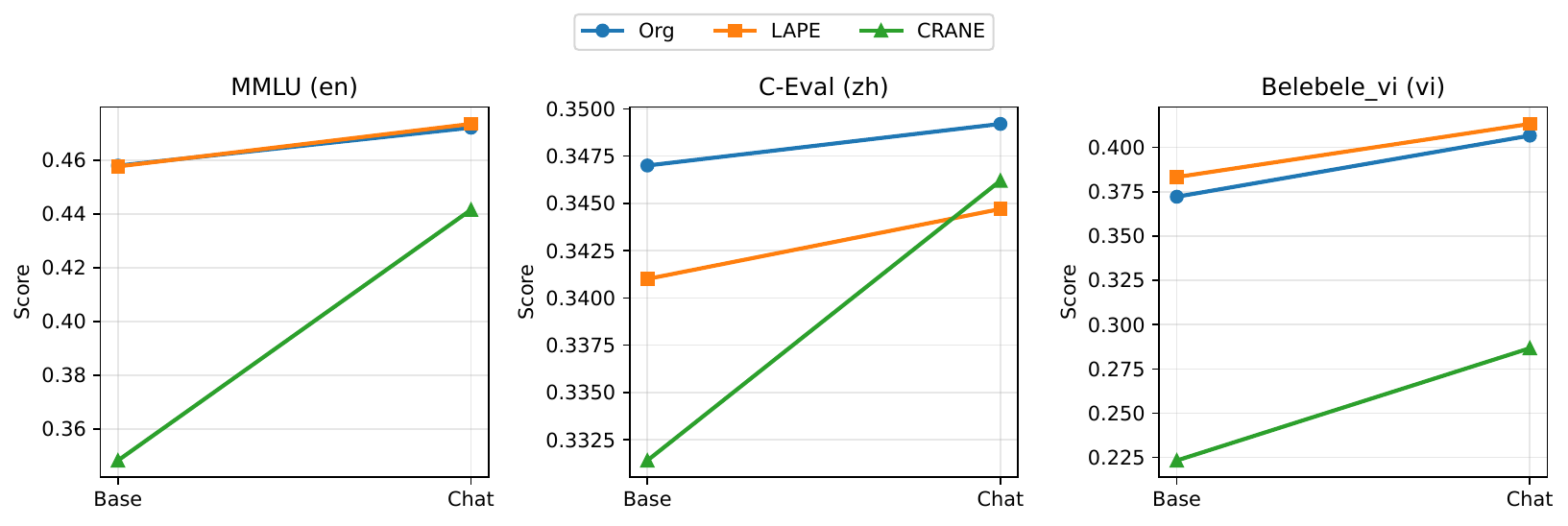}
  \caption{
  Raw target-language benchmark scores from Base to Chat under transferred neuron masks.
  Neuron sets are identified on \textsc{LLaMA2-7B-Base} and directly applied to \textsc{LLaMA2-7B-Chat}.
  Each panel corresponds to the target-language benchmark (MMLU / C-Eval / Belebele\_vi).
  }
  \label{fig:transfer_raw_scores}
\end{figure*}

Figure~\ref{fig:transfer_raw_scores} presents raw target-language benchmark scores under transferred
neuron masks, illustrating the absolute performance changes underlying the aggregated metrics.

\paragraph{Open-ended transfer.}
Table~\ref{tab:transfer_openended} reports open-ended generation results on the Chat model under
transferred neuron masks.
Consistent with NLU transfer, CRANE generally produces stronger target-language degradation than
activation-based baselines under comparable intervention budgets.
Given evaluation noise, these results are interpreted as supportive evidence of retained
functional influence.


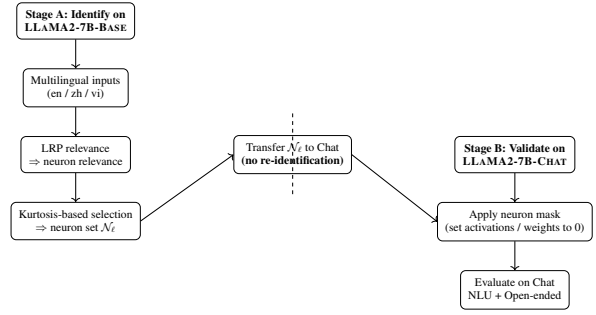
\begin{figure}[t]
\centering
\resizebox{\linewidth}{!}{%
\begin{tikzpicture}[
    node distance=8mm and 14mm,
    box/.style={
        draw,
        rounded corners,
        align=center,
        inner sep=6pt,
        font=\small
    },
    stage/.style={
        draw,
        rounded corners,
        align=center,
        inner sep=6pt,
        font=\bfseries\small
    },
    arrow/.style={->, thick}
]

\node[stage] (stageA) {Stage A: Identify on \\ \textsc{LLaMA2-7B-Base}};

\node[box, below=of stageA] (input) {Multilingual inputs \\ (en / zh / vi)};
\node[box, below=of input] (lrp) {LRP relevance \\ $\Rightarrow$ neuron relevance};
\node[box, below=of lrp] (kurt) {Kurtosis-based selection \\ $\Rightarrow$ neuron set $\mathcal{N}_\ell$};

\draw[arrow] (stageA) -- (input);
\draw[arrow] (input) -- (lrp);
\draw[arrow] (lrp) -- (kurt);

\node[box, right=28mm of lrp] (transfer) {
Transfer $\mathcal{N}_\ell$ to Chat \\
\textbf{(no re-identification)}
};

\draw[arrow] (kurt.east) -- (transfer.west);

\node[stage, right=28mm of transfer] (stageB) {Stage B: Validate on \\ \textsc{LLaMA2-7B-Chat}};

\node[box, below=of stageB] (mask) {
Apply neuron mask \\
(set activations / weights to 0)
};

\node[box, below=of mask] (eval) {
Evaluate on Chat \\
NLU + Open-ended
};

\draw[arrow] (stageB) -- (mask);
\draw[arrow] (mask) -- (eval);
\draw[arrow] (transfer.east) -- (mask.west);

\draw[dashed, thick] ($(transfer.north)+(0,6mm)$) -- ($(transfer.south)-(0,6mm)$);

\end{tikzpicture}
}

\caption{
\textbf{Transfer setting from Base to Chat.}
We first identify language-related neuron sets $\mathcal{N}_\ell$ on \textsc{LLaMA2-7B-Base}
using relevance attribution and kurtosis-based statistics.
The same neuron sets are then directly transferred and masked on
\textsc{LLaMA2-7B-Chat} \emph{without re-identification} to evaluate whether their
functional influence persists after post-training.
}
\label{fig:transfer-base-chat}
\end{figure}

\begin{table*}[t]
  \centering
  \small
  \setlength{\tabcolsep}{6pt}
  \resizebox{\textwidth}{!}{%
  \begin{tabular}{l c ccc ccc c}
    \toprule
    \multirow{2}{*}{Method} & \multirow{2}{*}{Mask} &
    \multicolumn{3}{c}{\textbf{Base model}} &
    \multicolumn{3}{c}{\textbf{Chat model}} &
    \multirow{2}{*}{LangSpec-F1} \\
    \cmidrule(lr){3-5}\cmidrule(lr){6-8}
    & & MMLU (en) & C-Eval (zh) & Belebele\_vi (vi)
      & MMLU (en) & C-Eval (zh) & Belebele\_vi (vi) & \\
    \midrule
    Org  & -- 
    & 0.4579 & 0.3470 & 0.3722 
    & 0.4720 & 0.3492 & 0.4067 
    & -- \\
    \midrule
    LAPE & en 
    & 0.4576 & 0.3351 & 0.3711 
    & 0.4734 & 0.3514 & 0.4044 
    & 0.0000 \\
    LAPE & vi 
    & 0.4553 & 0.3380 & 0.3833 
    & 0.4719 & 0.3477 & 0.4133 
    & 0.0000 \\
    LAPE & zh 
    & 0.4589 & 0.3410 & 0.3778 
    & 0.4719 & 0.3447 & 0.4078 
    & 0.0253 \\
    \midrule
    CRANE & en 
    & 0.3483 & 0.2801 & 0.2811 
    & 0.4415 & 0.3351 & 0.3367 
    & 0.1066 \\
    CRANE & vi 
    & 0.3517 & 0.2816 & 0.2233 
    & 0.4427 & 0.3276 & 0.2867 
    & 0.4316 \\
    CRANE & zh 
    & 0.4366 & 0.3314 & 0.3344 
    & 0.4645 & 0.3462 & 0.3733 
    & 0.0154 \\
    \bottomrule
  \end{tabular}
  }
  \caption{NLU transfer results. Neuron sets are identified on \textsc{LLaMA2-7B-Base} and directly applied (masked) on \textsc{LLaMA2-7B-Chat}. LangSpec-F1 is computed for the Chat model to summarize language-targeted degradation under masking.}
  \label{tab:transfer_nlu}
\end{table*}

\begin{table}[t]
  \centering
  \small
  \resizebox{\columnwidth}{!}{%
  \begin{tabular}{l c ccc c}
    \toprule
    Method & Mask & Open\_en & Open\_vi & Open\_zh & LangSpec-F1 \\
    \midrule
    Org  & -- 
    & 7.9857 & 7.0143 & 6.8286 
    & -- \\
    \midrule
    LAPE & en 
    & 7.9429 & 7.1286 & 6.9571 
    & 0.0107 \\
    LAPE & vi 
    & 7.6857 & 7.3000 & 6.9714 
    & 0.0000 \\
    LAPE & zh 
    & 8.0286 & 7.1857 & 6.7286 
    & 0.0289 \\
    \midrule
    CRANE & en 
    & 5.9429 & 3.2429 & 4.4429 
    & 0.2961 \\
    CRANE & vi 
    & 7.9143 & 4.9429 & 5.5143 
    & 0.3984 \\
    CRANE & zh 
    & 7.5714 & 6.1857 & 6.2000 
    & 0.1517 \\
    \bottomrule
  \end{tabular}%
  }
  \caption{Open-ended transfer results on \textsc{LLaMA2-7B-Chat}. We mask neuron sets identified from the Base model and evaluate on three language-specific open-ended benchmarks (scores in the 1--10 range).}
  \label{tab:transfer_openended}
\end{table}

\section{Conclusion}

We introduced \textbf{CRANE}, a relevance-based framework for analyzing
language-specific functional contributions in multilingual large language models.
By combining relevance attribution with neuron-level intervention, CRANE
operationalizes language specificity in terms of functional necessity rather
than activation-based correlation, and we further propose \textbf{LangSpec-F1}, a metric to quantify
language-selective functional effects under targeted neuron interventions.
Across multiple benchmarks and languages, CRANE consistently induces stronger
and more target-aligned degradation than activation-based baselines under matched
intervention budgets.
Under a strict transfer setting from a pretrained Base model to a post-trained
Chat model without re-identification, we find that a subset of Base-identified
neurons retains measurable functional influence after post-training, while others
shift.
More broadly, this work highlights the importance of distinguishing statistical
language correlation from functional language contribution at the neuron level,
and positions CRANE, together with the proposed metric, as a general framework
for studying multilingual representations and their evolution across training stages.

\section{Limitations}

CRANE currently relies on normalized kurtosis to characterize concentration
patterns in language-conditioned relevance distributions.
While effective and interpretable, kurtosis is only one possible statistic, and
other distributional measures may capture complementary aspects of
language-related structure.

Functional validation in this work is performed via neuron masking, which
constitutes a relatively coarse intervention.
More fine-grained causal techniques, such as neuron re-weighting or path-specific
analysis, may provide deeper insight into the mechanisms underlying
language-specific functional effects.

Our experiments focus on three languages and a single model family.
In addition, open-ended generation is evaluated using an LLM-as-a-judge protocol,
which may introduce variance.
Extending the analysis to broader language coverage, additional architectures,
and alternative evaluation paradigms remains important future work.

\section*{Ethics Statement}

This work analyzes internal mechanisms of multilingual language models and
evaluates their behavior on public benchmarks and curated prompts.
No personal or sensitive data are collected or used.

A potential risk is over-interpreting neuron-level findings as definitive causal
explanations of language behavior.
We mitigate this risk by grounding claims in controlled intervention, reporting
results cautiously, and clearly delineating the empirical scope and limitations
of our analysis.


\bibliography{main_0104_ARR}


\appendix

\end{document}